\documentclass[11pt,a4paper]{article}
\usepackage[hyperref]{acl2021}
\usepackage{times}
\usepackage{latexsym}

\usepackage{microtype}

\usepackage{multirow}
\usepackage{dingbat}
\usepackage{graphicx}
\usepackage[export]{adjustbox}
\usepackage[normalem]{ulem}
\useunder{\uline}{\ul}{}

\aclfinalcopy 

\title{End-to-End Speech Translation with Pre-trained Models and Adapters: UPC at IWSLT 2021}

\author{
  Gerard I. Gállego, Ioannis Tsiamas, Carlos Escolano,\\\textbf{José A. R. Fonollosa, Marta R. Costa-jussà} \\
  TALP Research Center, Universitat Politècnica de Catalunya, Barcelona \\
  \texttt{\{gerard.ion.gallego,ioannis.tsiamas,carlos.escolano}\\\texttt{jose.fonollosa,marta.ruiz\}@upc.edu} \\
}

\date{}

\begin{document}
\maketitle

\begin{abstract}
This paper describes the submission to the IWSLT 2021 offline speech translation task by the UPC Machine Translation group. The task consists of building a system capable of translating English audio recordings extracted from TED talks into German text. Submitted systems can be either cascade or end-to-end and use a custom or given segmentation. Our submission is an end-to-end speech translation system, which combines pre-trained models (Wav2Vec 2.0 and mBART) with coupling modules between the encoder and decoder, and uses an efficient fine-tuning technique, which trains only 20\% of its total parameters. We show that adding an Adapter to the system and pre-training it, can increase the convergence speed and the final result, with which we achieve a BLEU score of 27.3 on the MuST-C test set. Our final model is an ensemble that obtains 28.22 BLEU score on the same set. Our submission also uses a custom segmentation algorithm that employs pre-trained Wav2Vec 2.0 for identifying periods of untranscribable text and can bring improvements of 2.5 to 3 BLEU score on the IWSLT 2019 test set, as compared to the result with the given segmentation.

\end{abstract}

\section{Introduction}

    Typically, a speech translation (ST) system is composed of an automatic speech recognition (ASR) and a machine translation (MT) model, which is known as \textit{cascade} system. However, in recent years, \textit{end-to-end} models have gained popularity within the research community. These systems are encoder-decoder architectures capable of directly translating speech without intermediate symbolic representations. This approach solves classical shortcomings of \textit{cascade} ST systems, e.g. the error propagation or the slow inference time \cite{end2end-st}. Nevertheless, while there are plenty of data available to train ASR and MT systems, there are not as many datasets for ST, despite some recent efforts \cite{mustc,covost}. Moreover, this approach is inherently more difficult because the encoder has to perform both acoustic modeling and semantic encoding. For these reasons, end-to-end ST systems still struggle to achieve the performance of cascade ST models. Still, last year's IWSLT was the first time an end-to-end system had the best performance in the evaluation campaign \cite{best_iwslt2020,iwslt2020}. Hence, given the increasing interest in end-to-end ST systems, and the potential gains from advancing research on them, we decided to focus on developing such a system for this year's offline task.

    When there are not enough data for a task, a common practice is to use pre-trained components, like BERT \cite{bert} for various NLP tasks. In the ST field, the idea of pre-training the encoder for ASR was introduced by \citet{asr-pretrain} and has become a standard technique for developing modern end-to-end systems \cite{asr-pretrain-ex1,asr-pretrain-ex2}. By contrast, pre-training the decoder for MT does not lead to better performance \cite{no-mt-pretrain}. Recently, \citet{lna} proposed a multilingual ST system that combines a pre-trained Wav2Vec 2.0 \cite{wav2vec2.0} as the encoder and a pre-trained mBART decoder \cite{mBART}. Furthermore, they proposed a minimalist fine-tuning strategy that trains only the 20\% of the model parameters, while achieving similar performance to fine-tuning the whole model. From our perspective, this approach might become a turning point in the field, including bilingual scenarios like the IWSLT offline task. Hence, we decided to adopt this architecture\footnote{Since the pre-trained modules were trained on external data, our submission is unconstrained.} and fine-tuning strategy in our system (\S\ref{sec:pre-trained}). In addition, we introduce an Adapter module to extract better representations from the encoder (\S\ref{sec:adapteors}), and we propose a two-step training strategy (\S\ref{sec:exp_setup}) that brings improvements to the translation quality.

    During training, we used data augmentation techniques to boost our system's performance. Specifically, we applied randomized on-the-fly augmentations by adding an echo effect and modifying tempo and pitch (\S\ref{sec:data_aug}). Since our system works directly on the audio waveform, we could not use SpecAugment \cite{specaugment,specaugment-st}. Instead, we applied masking to the output of the Wav2Vec 2.0 feature extraction module, thereby obtaining a similar effect.

    The test data are provided with an automatic segmentation that does not ensure sentence-like segments. Considering the trend observed in 2019 and 2020 IWSLT offline task, where submission with own segmentation algorithms are strictly better than those with the given segmentation, we also decided to work with a custom segmentation algorithm. We base it on the approach of \citet{srpol2019}, but we replace the silence detection tool with an ASR system (\S\ref{sec:segm}). Our experiments on the IWSLT 2019 test set, show that our system works better when the data are segmented with our own segmentation algorithm (\S\ref{sec:results}).

\section{System description}
    We built an end-to-end ST system, mainly composed of pre-trained modules. We couple a Wav2Vec 2.0 encoder \cite{wav2vec2.0} and an mBART decoder \cite{mBART}, following the strategy proposed by \citet{lna}. When combining these two models, there is a length discrepancy between the target sentence length and the encoder output. For this reason, it is necessary to use a module to shorten the encoder output, which we refer to as the Length Adaptor.
    Additionally, we introduce an Adapter module to reduce the gap between the different modalities of the pre-trained models \cite{adapter}. A method that \citet{adapter-st} proved to be beneficial for ST models.
    
    \begin{figure}[t]
      \centering
      \includegraphics[width=\linewidth]{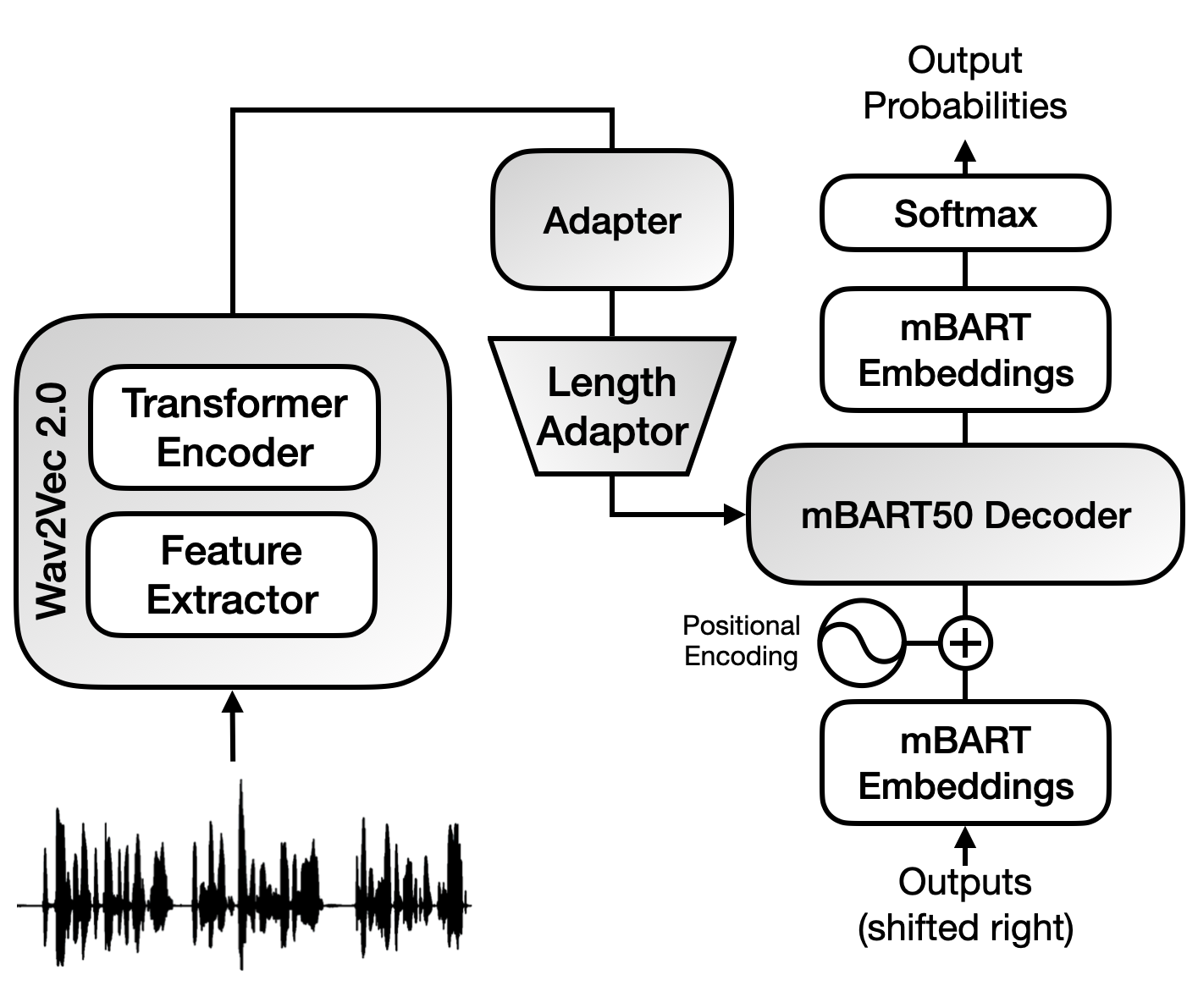}
      \caption{System overview. The original architecture proposed by \citet{lna} includes a pre-trained Wav2Vec 2.0 as the encoder, a pre-trained mBART decoder and a Length Adaptor. In this work, we add an Adapter module after the encoder.}
      \label{fig:system_overview}
    \end{figure}

    \subsection{Pre-trained modules} \label{sec:pre-trained}
        Our motivation is to get the most out of pre-trained components, which were obtained by self-supervision or supervised tasks. Concretely, we use a Wav2Vec 2.0 encoder and an mBART decoder, both trained initially by self-supervision and fine-tuned for ASR and multilingual MT, respectively.
        
        \textbf{Wav2Vec 2.0} is a speech encoder proposed by \citet{wav2vec2.0}. This model is pre-trained by self-supervision, i.e. without explicit targets such as transcriptions. Its main contribution is that it achieves excellent performance in ASR after fine-tuning it with just a few minutes of transcribed speech. Moreover, it can process raw audio waveforms directly, unlike other systems which work with spectrogram-like representations \cite{s-transformer}.
        
        This model is composed of two main blocks. Firstly, a feature extractor made of seven 1-D convolutional layers processes the raw audio waveform. The representation obtained from this step has a stride of 20ms between samples, and each one has a receptive field of 25ms. Secondly, a Transformer \cite{transformer} encoder with 24 layers extracts contextualized representations. For the purpose of our system, we discard the rest of the components that are used during the self-supervised pre-training (e.g. the quantization modules).
        
        The Wav2Vec 2.0 model that we employ is already fine-tuned on ASR. Specifically, we use the \textit{Large} architecture, pre-trained with 53.2k hours of untranscribed speech from LibriVox \cite{librilight}, fine-tuned on the 960h of transcribed speech from Librispeech \cite{librispeech}, and on pseudo-labels \cite{wav2vec-selftraining}.
        
        \textbf{mBART} is a sequence-to-sequence denoising autoencoder, which reconstructs the input text sentence given a corrupted version of it \cite{mBART}. It follows the same approach as BART \cite{BART} but, instead of using just English monolingual data, it is trained with multiple languages. This strategy does not require any parallel corpora, so it can be used as a pre-training step and then fine-tuned for MT tasks.

        Specifically, we use the 12-layer Transformer decoder of an mBART model, fine-tuned on multilingual MT, from English to 49 languages \cite{mBART50}.
    
    \subsection{Coupling modules} \label{sec:adapteors}
        In addition to the two main blocks that constitute our system, we implement another two other modules placed after the Wav2Vec 2.0 encoder (Figure \ref{fig:system_overview}). The objective of these modules is to overcome the multimodal gap by adapting the encoder output to the decoder. With them, we adapt the representations to the decoder's modality, and reduce its length.
        
\begin{figure}[t]
          \centering
          \includegraphics[width=0.85\linewidth]{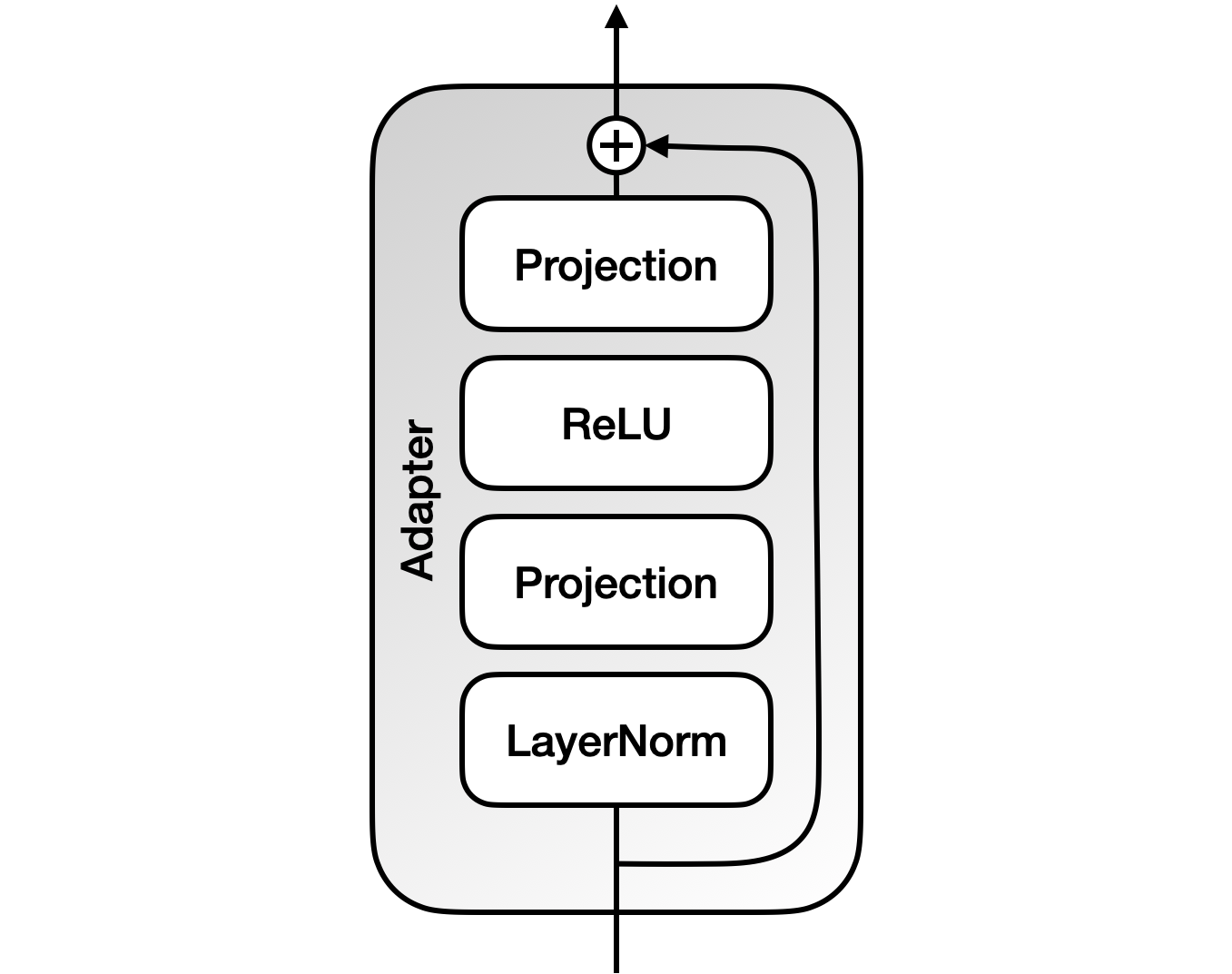}
          \caption{Adapter module}
          \label{fig:adapter}
        \end{figure}

        The \textbf{Adapter} is a module that was introduced by \citet{adapter} to adapt pre-trained models to multiple tasks. The Adapter projects its input to a higher-dimensional space before reducing it to the original size. Moreover, it applies layer normalization at the input \cite{layernorm}, a ReLU activation after the first projection and a residual connection (Figure \ref{fig:adapter}).

        In work done by \citet{adapter-st}, they proposed to use this module to adjust the representation from the speech encoder to the language-specific decoders. Hence, we have used this module with a similar purpose, since we also needed to combine different pre-trained components and modalities.
                
        The \textbf{Length Adaptor} is a module that reduces the length discrepancy between the input and output sequences. It achieves an 8x down-sampling of the encoder representation by applying a stack of 3 convolutional layers with a kernel size of 3 and a stride of 2.
    
    \subsection{LNA Finetuning} \label{sec:lna_ft}
        We follow the LayerNorm and Attention (LNA) fine-tuning strategy proposed by \citet{lna}. The main idea is that only some of the modules of Wav2Vec 2.0 and mBART need to be fine-tuned to build a system capable of ST. More specifically, these are the layer normalization, encoder self-attention and encoder-decoder attention, which account for the 20\% of the total parameters. It was shown that this minimal fine-tuning not only creates a powerful ST system, but its performance also approximates what is obtained by fine-tuning all the parameters. Even more importantly, it allows fast and memory-efficient training, which enabled us to work with such a large architecture.

\section{Data}

    Here we introduce the datasets used for our experiments and describe the filtering and data augmentation methods that were employed during training.

    \subsection{Datasets} \label{sec:datasets}
    
        For our experiments, we are using the English-to-German data from three ST datasets, namely the MuST-C v2 \footnote{The second version of MuST-C has not been officially released yet, but the En-De data is available in advance at \url{https://ict.fbk.eu/must-c/.}} \cite{mustc}, EuroparlST \cite{europarlst} and CoVoST 2 \cite{covost} \footnote{The EuroparlST and CoVoST 2 data are converted to 16khz, which is required for the input of the Wav2Vec 2.0 encoder.}. Our training set is a concatenation of the respective train splits of these datasets, while we discarded the train-noisy split of EuroparlST due to low quality. We only consider MuST-C to be in-domain, since its data come from TED talks, and thus EuroparlST and CoVoST are considered out-of-domain due to differences in setting, use of language and segment duration. Given this, our development data are comprised only of the development split of MuST-C, which allows us to concatenate the development splits of EuroparlST and CoVoST to our training data. Furthermore, we down-sample the CoVoST splits during each training epoch to shift the importance towards the MuST-C data. We do not down-sample EuroparlST due to its already small size compared to MuST-C (Table \ref{tab:train_data}). We use two different sets for evaluating the performance of our system, the test split of MuST-C and the IWSLT 2019 test set \cite{iwslt2019}. The latter one provides us with an opportunity to additionally test our segmentation algorithm, since the given segmentation and the reference translations are not perfectly aligned nor sentence-like. Finally we generate our predictions for the IWSLT test sets of 2020 \cite{iwslt2020} and 2021 \cite{iwslt2021}, for which the reference translations have not been made available (Table \ref{tab:dev_test_data}). We do not use the rest of the IWSLT test sets, since they are already included in the 2nd version of MuST-C.
        
        \begin{table}[t]
            \centering
            \begin{tabular}{lccc}
                \hline
                Split & \begin{tabular}[c]{@{}c@{}}Available\\ References\end{tabular} & \begin{tabular}[c]{@{}c@{}}Aligned\\ Segmentation\end{tabular} \\ \hline
                MuST-C-dev    & \checkmark & \checkmark \\
                MuST-C-test   & \checkmark & \checkmark \\
                IWSLT.tst2019 & \checkmark &            \\
                IWSLT.tst2020 &            &            \\
                IWSLT.tst2021 &            &            \\ \hline
            \end{tabular}
            \caption{Development and Test splits}
            \label{tab:dev_test_data}
        \end{table}

    \subsection{Data filtering} \label{sec:filtering}
        
        We remove examples where the duration of the source audio is more than 25 seconds (400,000 samples) to avoid out-of-memory errors during the training of the ST system. Apart from that, we use another two filtering stages to ensure that our training data are of high quality, for which we provide the details bellow. The size of the training data after all the filtering stages can be found in Table \ref{tab:train_data}.
        
        \paragraph{Text Filtering.} We perform text filtering on the target German text of MuST-C to remove speaker names and non-textual events. Speaker names in MuST-C are used to differentiate between speakers, when multiple of them are interacting in a talk. They appear in the beginning of a sentence, as full names or capitalized initials, followed by a colon. We remove the text in the beginning of each sentence if it matches the described pattern. Non-textual events are enclosed in parentheses, with some common examples being “(Gelächter)” or “(Applaus)”, which are the German translations of “laughter” and “applause”. In such cases we keep the examples but we remove the events. The only exception are cases where there are actual utterances coming from a secondary speaker. For those cases, we strip the parentheses and the speaker names. For EuroparlST, large numbers use spaces as the thousands-separator, which we convert to commas, in order to match the number format of MuST-C and IWSLT data. No specific text filtering is done for CoVoST. Finally, we remove the examples that are empty after applying the text filtering.
        
        \paragraph{ASR Filtering.} For the final stage of filtering, we use an Automatic Speech Recognition (ASR) model to identify noisy examples. We employ a pre-trained Wav2Vec 2.0 \cite{wav2vec2.0}, from the HuggingFace Transformers library \cite{transformers} and perform inference on all our training examples. The pre-trained Wav2Vec 2.0 is quite effective in this task and achieves an average word-error-rate (WER) of 0.135. Consecutively we remove those examples where the predicted text has a WER greater than 0.5, as compared to its English reference text. At this stage of filtering we remove approximately $4\%$ of our total training data. For ASR inference, all English target text was normalized, lower-cased, stripped from punctuation and numbers were converted to spelled-out words. 

        \begin{table}[t]
            \centering
            \resizebox{\columnwidth}{!}{
                \begin{tabular}{l|cc|c}
                \hline
                    Split            & Original & Filtered & S.Ratio \\ \hline
                    MuST-C-train     & 450      & 415      & 1.0          \\
                    EuroparlST-train & 77       & 75       & 1.0          \\
                    EuroparlST-dev   & 3        & 3        & 1.0          \\
                    CoVoST-train     & 430      & 410      & 0.3          \\
                    CoVoST-dev       & 26       & 24       & 0.3          \\ \hline
                    Total            & 986      & 927      & -        
                \end{tabular}
            }
            \caption{Training splits with their original and filtered sizes measured in hours, and the sampling ratios for each split in every training epoch.}
            \label{tab:train_data}
        \end{table}

    \subsection{Data augmentation} \label{sec:data_aug}
    
        Data augmentation has been shown to provide increased performance in both ASR \cite{specaugment} and ST \cite{fbk2019}, by enriching and diversifying the training data. Thus, following \citet{srpol2019}, we perform data augmentation on the English source audio. We apply the “tempo” and “pitch” effects to force our system to adapt to speeches of different speeds, and the “echo” effect to simulate the echoing which is usually present in large rooms, where TED talks are taking place. Compared to \citet{srpol2019}, we replace the “speed” effect in favor of “pitch”, since “speed” also modifies the “tempo”, which is a separate effect. Data augmentation is applied on-the-fly, during training, using WavAugment \cite{wavaugment2020}, which is build on top of the SoX library \footnote{SoX - \url{https://sox.sourceforge.net}}. Each example in the batch has a probability of $p_{aug} = 0.8$ to be augmented, in which case we apply all three effects to it. We sample uniformly the parameters of each effect from the ranges shown at Table \ref{tab:aug}.
        
        \begin{table}[t]
            \centering
            \begin{tabular}{lcc}
                \hline
                Parameter  & Min value & Max value \\ \hline
                tempo      & 0.85      & 1.3       \\
                pitch      & -300      & 300       \\
                echo-delay & 20        & 200       \\
                echo-decay & 0.05      & 0.2       \\ \hline
            \end{tabular}
            \caption{Data Augmentation parameter ranges. Echo is controlled by two parameters. Tempo and echo-decay are coefficients, pitch is measured in semitones and echo-delay in milliseconds.}
            \label{tab:aug}
        \end{table}

    \subsection{Data Segmentation}
    \label{sec:segm}
\begin{figure}[t]
            \centering
            \includegraphics[width=\columnwidth]{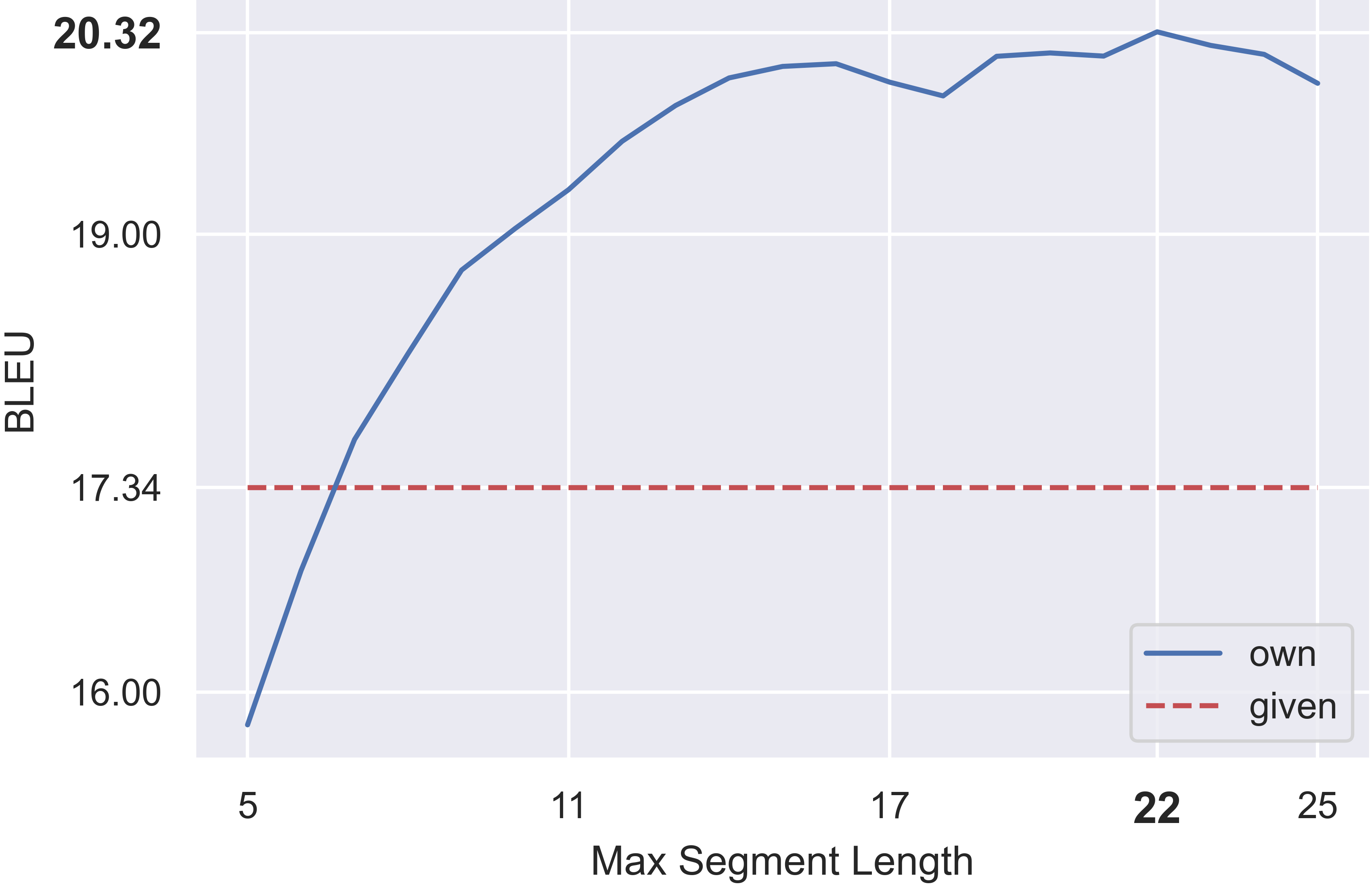}
            \caption{BLEU scores for our segmentation algorithm with different values of $max\_seg\_len$ on IWSLT.tst2019. X-axis is in seconds. With red color is the BLEU score for the given segmentation.}
            \label{fig:segm}
        \end{figure}

        Similarly to 2019 and 2020 \cite{iwslt2019,iwslt2020}, this year's evaluation data are segmented using an automatic tool \cite{lium}, which does not ensure that segments are proper sentences nor that they are aligned with the translated text. This assigns extra importance to developing methods for proper segmentation of the audio data, which was confirmed in the previous year's evaluation campaign, where all top submissions used their own segmentation algorithm. For creating our own segmentation of the IWSLT 2020 and 2021 test sets, we modify the technique described in \citet{srpol2019}, where they use a silence detection tool \footnote{Audacity - \url{https://www.audacityteam.org}} to progressively split each audio file into smaller segments. Their algorithm terminates when all segments do not exceed a maximum segment length ($max\_seg\_len$) threshold, which they tune to maximize the BLEU score on IWSLT 2015 test set \cite{iwslt2015}. In our approach we replace the silence detection tool with a pre-trained Wav2Vec 2.0 model \cite{wav2vec2.0} from the Huggingface Transformer library \cite{transformers}, to identify periods of untranscribable English text. Since the IWSLT 2015 test set is included in MuST-C v2, we tune our algorithm on IWSLT 2019 test set. First, we perform inference with Wav2Vec 2.0 on the IWSLT 2019 test set, and obtain a token prediction for every 20ms for each audio file. Then we proceed to split each audio file on the largest untranscribable period, which is identified by the absence of English characters in it. The algorithm terminates when the max segment length condition is satisfied or no further splits are possible due to a minimum untranscribable period length, which we set to 0.2 seconds. We test $max\_seg\_len \in [5,25]$, and for each value we produce a segmentation, generate translations using one of our ST systems \footnote{For the purpose of this experiment we used the best checkpoint from the LNA-ED-Adapt experiment (Table \ref{tab:results})}, use the mwerSegmenter \footnote{\url{https://github.com/jniehues-kit/SLT.KIT}} software to align the generated translations with the reference translations, and finally obtain a BLEU score using SACREBLEU \cite{sacrebleu}. We find that the maximum BLEU score is obtained using  $max\_seg\_len = 22$ seconds (Figure \ref{fig:segm}), which we use to segment the IWSLT 2020 and 2021 test sets for our submission.
        
\section{Experiments}

    Here we describe our experiments, along with their implementation details and the results on MuST-C and the IWSLT 2019 test set.

    \subsection{Experimental Setup}
    \label{sec:exp_setup}
    
        \paragraph{LNA-ED} The first experiment is to train our baseline model, which is an encoder-decoder model with a length adaptor module (\S \ref{sec:adapteors}) in between. As in \citet{lna}, we initialize the encoder with a pre-trained Wav2Vec 2.0, the decoder with the decoder of a pre-trained mBART50 (\S \ref{sec:pre-trained}) and we only train the parameters of the layer normalization in both encoder and decoder, the encoder self-attention in the encoder, the encoder cross-attention in the decoder, and Length Adaptor (\S \ref{sec:lna_ft}).
        
        \paragraph{LNA-ED-Adapt} Following we experiment with adding an Adapter module (\S \ref{sec:adapteors}) prior to the Length Adaptor, while we train the same parameters as in LNA-ED. We expect that this module will adapt the encoder output to the decoder's modality, before down-sampling it with the convolutional layers of the Length Adaptor.
        
        \paragraph{LNA-ED-Adapt-2step} Our next experiment aims at initializing all the sub-modules from pre-trained checkpoints. Thus, our first step is to train only the coupling modules of the LNA-ED-Adapt system, while everything else is frozen. Then, in the second step we proceed by training all the active parameters of LNA-ED-Adapt. We hypothesize that in the prior experiments the initially random weights of the coupling modules are slowing down the learning process and potentially also hurting the final performance of the system.
        
        \paragraph{In-domain FT} We experiment with fine-tuning our systems for some additional epochs only on the in-domain data of MuST-C. During this fine-tuning we also disable data augmentation.
        
        \paragraph{Ckpt AVG} We average checkpoints around the best, indicated by the highest BLEU score in the development split of MuST-C. This technique has been shown to provide more generalizable models, achieving higher scores in the hidden test sets \cite{fbk2020,bhanss2020}.
        
        \paragraph{Ensemble} For our final model, we ensemble our two best single models. To increase the diversity of the two single models and, consecutively, the performance of the ensemble, we choose one that is further fine-tuned on in-domain data and one that is not.
        We expect that, although there is a potential boost in the performance of a system by fine-tuning to in-domain data, there is the risk of catastrophic forgetting of the more general data properties of the combined and augmented corpus. Thus, we combine a model specialized to the in-domain data and one which is potentially more general.

    \subsection{Implementation details}

        For the encoder and decoder of our models, we are using the same architecture as the Wav2Vec 2.0 and mBART decoder (\S \ref{sec:pre-trained}). More specifically the encoder has a 7-layer convolutional feature extractor and a 24-layer Transformer encoder, while the decoder has 12 layers. The feature extractor has 512 channels, while each Transformer layer has a dimensionality of 1024, feed-forward dimension of 4096, and 16 heads. For the Adapter, we use an inner dimensionality of 4096, which was shown to work better in \citet{adapter-st} and for the Length Adaptor we set the kernel size to 3 and the stride to 2. The decoder uses a vocabulary of 250,000 tokens, and the embedding layer is shared between source and target.
        
        We train all our models with the LNA method (\S \ref{sec:lna_ft}), unless stated otherwise. The training data for each epoch are coming from the 5 splits show in  Table \ref{tab:train_data}, with their respective sampling ratios. We limit the length of the source examples to 400,000 samples (i.e. 25 seconds) and to 1024 tokens for the target. For each example, we apply data augmentation (\S \ref{sec:data_aug}) on the source speech and subsequently, normalize it to zero mean and unit variance. We construct mini-batches with a maximum of 440,000 samples, and use data parallelism on 4 GPUs and gradient accumulation with 16 steps, to increase the effective batch size by a factor of 64.
        
        For optimization we use Adam \cite{adam} with parameters $\beta_1 = 0.99$, $\beta_2 = 0.98$. We set the base learning rate to $10^{-4}$, which is controlled during training by a tri-stage scheduler with the ratios for the warm-up, hold and decay phases being 0.15, 0.15, and 0.7 accordingly, and initial and final scales of 0.01. We clip gradients to a maximum norm of 20, and we apply a dropout of 0.1 before every non-frozen layer or sub-layer in our models. Following \citet{mBART25}, the optimizer is minimizing the standard cross-entropy loss with a label smoothing of 0.2. All models are trained for 16 epochs (approximately 23,000 updates), apart from the pre-training step of the LNA-ED-Adapt-2step and the in-domain fine-tuning, which are carried out for 4 epochs.
        
        We pick the checkpoint with the highest BLEU score on the development set of MuST-C, for which then we report the BLEU on the test set of MuST-C and the IWSLT 2019 test set. We ensemble the 2 best models according to the BLEU score on the test set of MuST-C. For generation, we are using a standard beam search with a size of 5. All our experiments are done in a machine with 4 Nvidia GeForce RTX 2080 Ti GPUs, using 16 floating-point precision, and are implemented in fairseq \cite{wang-etal-2020-fairseq}. The training of each model took approximately 60 hours. The code for our experiments is available in a public repository\footnote{\url{https://github.com/mt-upc/iwslt-2021}}.
        
    \subsection{Results}
        \label{sec:results}
\begin{figure}[t]
            \centering
            \includegraphics[width=\columnwidth]{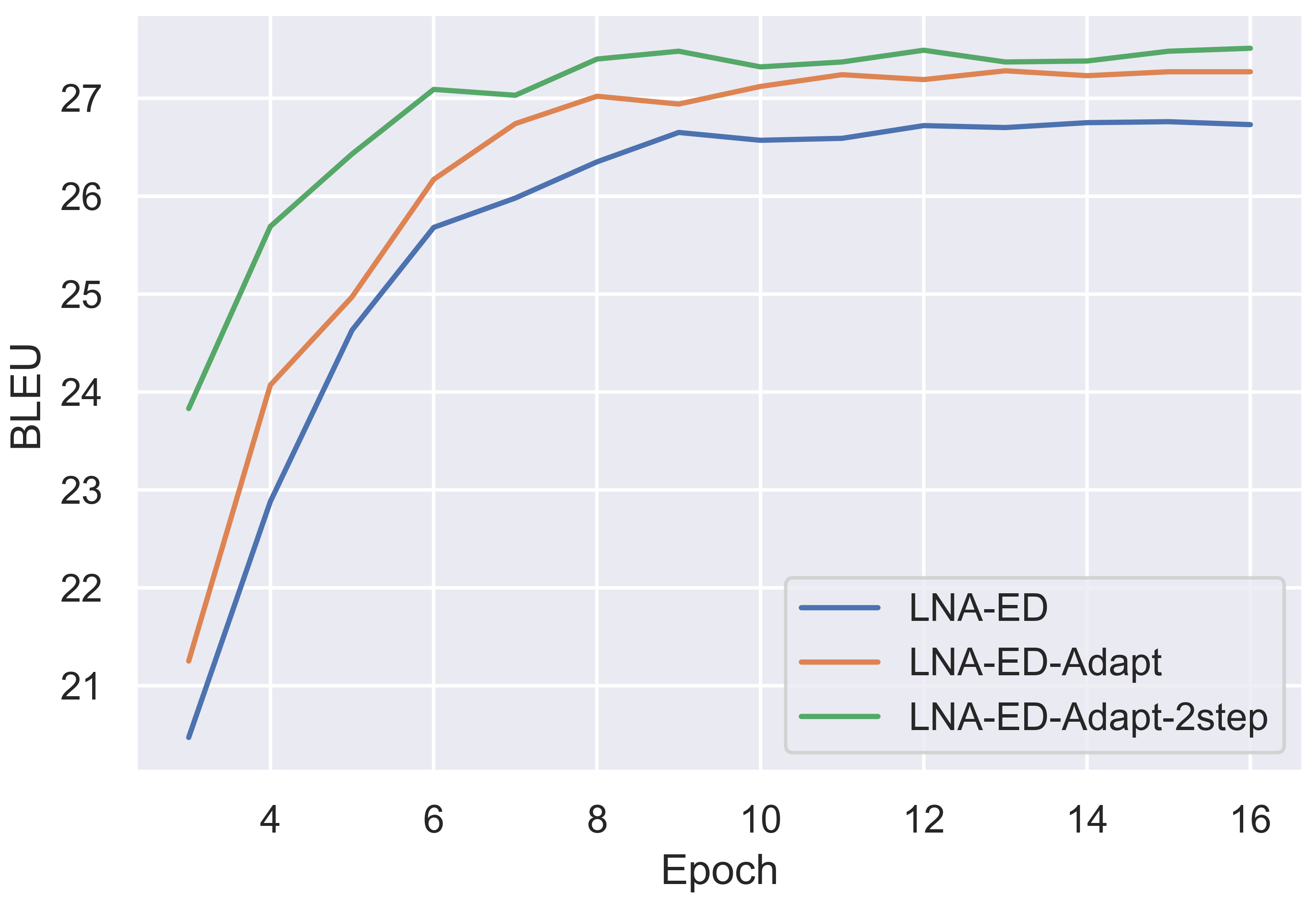}
            \caption{BLEU scores on MuST-C-dev during training}
            \label{fig:learning_curve}
        \end{figure}

        The results of our experiments (\S\ref{sec:exp_setup}) on the development and test sets of MuST-C can be found in Table \ref{tab:results}. We also provide the BLEU score on the IWSLT 2019 test set, for both the given and our own segmentation, using a max segment length of 22 (\S \ref{sec:segm}). The addition of the Adapter module provides an increase of 0.76 BLEU in MuST-C test set, as compared to LNA-ED. We observe that training our system in two steps can bring further improvements to the quality of translations. The first step of training of the LNA-ED-Adapt-2step experiment, with only the coupling modules being active, achieves a BLEU score of 15.54 after 4 epochs of training. Subsequently, the 2nd step is initialized from a much better checkpoint, as compared to the previous experiments, and can converge faster, as we can observe in Figure \ref{fig:learning_curve}, eventually achieving a BLEU score of 27.25.
        
        Both the LNA-ED-Adapt and LNA-ED-Adapt-2step bring improvements to the base model, without a significant computational burden. The Adapter module has 8.4 million parameters, which accounts for an increase of only $5\%$ in the total trainable parameters of the LNA method. In the first step of LNA-ED-Adapt-2step we are only training 9.1 million parameters for 4 epochs, a process that is completed rather fast compared to the training of the second step.
        
        We achieve increased performance by fine-tuning the best checkpoint of LNA-ED-Adapt on the in-domain data of MuST-C for another 4 epochs. What stands out from this further fine-tuning is the large improvements in the IWSLT 2019 test set, providing us with our best score on the own segmentation from a single model. Due to time constraints, we carried out this fine-tuning only on LNA-ED-Adapt and not on LNA-ED-Adapt-2step. Finally, we average the checkpoints around the best for the in-domain fine-tuned LNA-ED-Adapt and the LNA-ED-Adapt-2step. Using them in an ensemble, we obtain a BLEU score of 28.22 on the test set of MuST-C, which is an improvement of 0.92 points from our best single model, while smaller improvements are observed in the IWSLT 2019 test set.
        
        Regarding the translation quality on the IWSLT 2019 test set, we can observe that using our own segmentation algorithm, we can obtain large improvements, from 2.5 to 3 in BLEU score.
        
        \begin{table}[t]
            \centering
            \resizebox{\columnwidth}{!}{
            \begin{tabular}{l|cc|cc}
            \hline
            Model                          & \multicolumn{2}{c|}{MuST-C}                & \multicolumn{2}{c|}{IWSLT.tst2019}          \\
                                           & dev                 & test                 & given                & own                  \\ \hline
            LNA-ED                         & 26.76               & 26.23                & 17.25                & 20.06                \\ \hline
            LNA-ED-Adapt                   & 27.28               & 26.99                & 17.34                & 20.32                \\
            $\hookrightarrow$ In-domain FT & 27.36               & 27.25                & 18.79                & \textbf{21.29}       \\
            $\hookrightarrow$ ckpt AVG (a) & 27.36               & 27.29                & \textbf{18.97}       & 21.13                \\ \hline
            LNA-ED-Adapt-2step             & 27.49               & 27.25                & 17.56                & 20.37                \\
            $\hookrightarrow$ ckpt AVG (b) & \textbf{27.5}       & \textbf{27.3}        & 17.51                & 20.38                \\ \hline
            Ensemble (a) $\&$ (b)          & {\ul \textbf{28.5}} & {\ul \textbf{28.22}} & {\ul \textbf{19.05}} & {\ul \textbf{21.43}} \\ \hline
            \end{tabular}
            }
            \caption{BLEU scores on dev and test sets of MuST-C and on the IWSLT.tst2019 with given and own segmentation. With bold are the best scores by single models and with underlined bold are the best scores overall.}
            \label{tab:results}
        \end{table}
        
    \subsection{Submission results}
        \label{sec:final_results}

    \begin{table}[h!]
        \centering
        \resizebox{\columnwidth}{!}{
            \begin{tabular}{cc|c|ccc}
            \hline
            \multirow{2}{*}{Model} & \multirow{2}{*}{Segmentation} & \multicolumn{4}{c|}{Reference} \\
                                   &                               & 2020       & 2021$\dagger$    & 2021$\ddagger$ & 2021$\star$  \\
            \hline
            \textbf{Ensemble}      & \textbf{Own}                  & \textbf{24.6} & \textbf{21.8}     & \textbf{18.3}   & \textbf{30.6}  \\
            Ensemble               & Given                         & 20.5          & 19.5              & 16.0            & 26.7  \\
            Single                 & Own                           & 23.0          & 20.7              & 17.5            & 29.0  \\
            Single                 & Given                         & 19.0          & 18.4              & 15.0            & 25.0  \\
            \hline
            \end{tabular}}
        \caption{Final results of our submission on the IWSLT 2020 and 2021 test sets, measured in BLEU, against the IWSLT ($\dagger$) and TED ($\ddagger$) references separately and both at once ($\star$). With bold is our primary submission. The \textit{Single} is our best single model from Table \ref{tab:results} (LNA-ED-Adapt-2step with ckpt AVG) and the \textit{Ensemble} to the ensemble of our best single model and the LNA-ED-Adapt with In-domain FT and ckpt AVG.}
        \label{tab:final_results_2021}
    \end{table}

        There are two references available for this year's test set \cite{iwslt2021}, one corresponding to the official TED talks subtitles and another generated by the IWSLT organizers. Our primary submission is the ensemble of the two best models with our segmentation, which scores 18.3 BLEU against the TED references, 21.8 BLEU with the IWSLT references, and 30.6 BLEU with both together (Table \ref{tab:final_results_2021}). Meanwhile, when using the given segmentation, we get a decrease of 2.3 BLEU in both references, which is consistent to the results obtained in the IWSLT 2019 test set (Table \ref{tab:results}). As a contrastive system, we also submitted the results obtained with our best single model, corresponding to the LNA-ED-Adapt-2step model with checkpoint averaging. This system scores approximately 1 BLEU less with respect to the ensemble, similarly to the results we get in the IWSLT 2019 test set (Table \ref{tab:results}).
        
        We also evaluated our systems on the IWSLT 2020 test set, for tracking year-to-year progress. Our best model obtains a BLEU score of 24.6 (Table \ref{tab:final_results_2021}) and, in general, the results follow the same trend as on the IWSLT 2021 test set. For comparison, our best model would have been place 3rd in last year's leaderboard \cite{iwslt2020}, 0.7 BLEU points behind the best system \cite{best_iwslt2020}.

\section{Conclusions}
    We described the UPC Machine Translation group participation in the IWSLT 2021 offline ST task. We built our system by combining pre-trained components, using Wav2Vec 2.0 as an encoder and an mBART decoder. In order to fine-tune such a large model with approximately 770 million parameters, we followed the strategy proposed by \citet{lna}, in which just a 20\% of the parameters are trained. Originally, this method was proposed for multilingual ST, and it had not been applied to initialize a bilingual system yet. With this approach, we got a score of 26.23 BLEU in the MuST-C test set. Then, we introduced an Adapter module to reduce the gap between the different modalities of the pre-trained components, which brought an improvement of 0.76 BLEU. We also explored a two-step training where we initialized the coupling modules before fine-tuning the rest of the model, which resulted in an increase of 1.02 BLEU with respect to the original model. Furthermore, we applied other techniques like fine-tuning with in-domain data, checkpoint averaging and ensembling our two best models. Our final score in the MuST-C test set was 28.22 BLEU. Apart from using Wav2Vec 2.0 as the encoder of our ST system, we additionally leveraged it in our ASR-based data filtering and as part of our segmentation algorithm. Applying this custom segmentation we gained an increase of 2.5 to 3 BLEU score in the IWSLT 2019 test set, as compared to the result of with given segmentation.
    
    As was shown in \citet{lna}, and confirmed in this work for a bilingual scenario, large pre-trained models can be very effective in ST. We believe that future work should focus on exploring better methods to adapt these pre-trained models to new languages and tasks, with Adapter modules being promising candidates.

\section*{Acknowledgments}

This work was supported by the project ADAVOICE, PID2019-107579RB-I00 / AEI / 10.13039/501100011033 and by the European Research Council (ERC) under the European Union’s Horizon 2020 research and innovation programme (grant agreement No. 947657).

\bibliographystyle{acl_natbib}
\bibliography{anthology,acl2021}

\begin{thebibliography}{40}
\expandafter\ifx\csname natexlab\endcsname\relax\def\natexlab#1{#1}\fi

\bibitem[{Anastasopoulos et~al.(2021)Anastasopoulos, Bojar, Bremerman, Cattoni,
  Elbayad, Federico, Ma, Nakamura, Negri, Niehues, Pino, Salesky, St\"{u}ker,
  Sudoh, Turchi, Waibel, Wang, and Wiesner}]{iwslt2021}
Antonios Anastasopoulos, Ondřej Bojar, Jacob Bremerman, Roldano Cattoni, Maha
  Elbayad, Marcello Federico, Xutai Ma, Satoshi Nakamura, Matteo Negri, Jan
  Niehues, Juan Pino, Elizabeth Salesky, Sebastian St\"{u}ker, Katsuhito Sudoh,
  Marco Turchi, Alex Waibel, Changhan Wang, and Matthew Wiesner. 2021.
\newblock {Findings of the IWSLT 2021 Evaluation Campaign}.
\newblock In \emph{Proceedings of the 18th International Conference on Spoken
  Language Translation (IWSLT 2021)}, Online.

\bibitem[{Ansari et~al.(2020)Ansari, Axelrod, Bach, Bojar, Cattoni, Dalvi,
  Durrani, Federico, Federmann, Gu, Huang, Knight, Ma, Nagesh, Negri, Niehues,
  Pino, Salesky, Shi, St{\"{u}}ker, Turchi, Waibel, and Wang}]{iwslt2020}
Ebrahim Ansari, Amittai Axelrod, Nguyen Bach, Ondrej Bojar, Roldano Cattoni,
  Fahim Dalvi, Nadir Durrani, Marcello Federico, Christian Federmann, Jiatao
  Gu, Fei Huang, Kevin Knight, Xutai Ma, Ajay Nagesh, Matteo Negri, Jan
  Niehues, Juan Pino, Elizabeth Salesky, Xing Shi, Sebastian St{\"{u}}ker,
  Marco Turchi, Alexander~H. Waibel, and Changhan Wang. 2020.
\newblock \href {https://doi.org/10.18653/v1/2020.iwslt-1.1} {{FINDINGS} {OF}
  {THE} {IWSLT} 2020 {EVALUATION} {CAMPAIGN}}.
\newblock In \emph{Proceedings of the 17th International Conference on Spoken
  Language Translation, {IWSLT} 2020, Online, July 9 - 10, 2020}, pages 1--34.
  Association for Computational Linguistics.

\bibitem[{Ba et~al.(2016)Ba, Kiros, and Hinton}]{layernorm}
Jimmy~Lei Ba, Jamie~Ryan Kiros, and Geoffrey~E Hinton. 2016.
\newblock Layer normalization.
\newblock \emph{arXiv preprint arXiv:1607.06450}.

\bibitem[{Baevski et~al.(2020)Baevski, Zhou, Mohamed, and Auli}]{wav2vec2.0}
Alexei Baevski, Yuhao Zhou, Abdelrahman Mohamed, and Michael Auli. 2020.
\newblock \href
  {https://proceedings.neurips.cc/paper/2020/file/92d1e1eb1cd6f9fba3227870bb6d7f07-Paper.pdf}
  {wav2vec 2.0: A framework for self-supervised learning of speech
  representations}.
\newblock In \emph{Advances in Neural Information Processing Systems},
  volume~33, pages 12449--12460. Curran Associates, Inc.

\bibitem[{Bahar et~al.(2019)Bahar, Zeyer, Schlüter, and Ney}]{specaugment-st}
Parnia Bahar, Albert Zeyer, Ralf Schlüter, and Hermann Ney. 2019.
\newblock \href {https://doi.org/10.5281/ZENODO.3525010} {On {Using}
  {SpecAugment} for {End}-to-{End} {Speech} {Translation}}.
\newblock In \emph{Proceedings of the 16th {International} {Workshop} on
  {Spoken} {Language} {Translation}}, Hong Kong.
\newblock Publisher: Zenodo.

\bibitem[{Bansal et~al.(2019)Bansal, Kamper, Livescu, Lopez, and
  Goldwater}]{no-mt-pretrain}
Sameer Bansal, Herman Kamper, Karen Livescu, Adam Lopez, and Sharon Goldwater.
  2019.
\newblock \href {https://doi.org/10.18653/v1/N19-1006} {Pre-training on
  high-resource speech recognition improves low-resource speech-to-text
  translation}.
\newblock In \emph{Proceedings of the 2019 {Conference} of the {North}
  {American} {Chapter} of the {Association} for {Computational} {Linguistics}:
  {Human} {Language} {Technologies}}, pages 58--68, Minneapolis, Minnesota.
  Association for Computational Linguistics.

\bibitem[{Bapna and Firat(2019)}]{adapter}
Ankur Bapna and Orhan Firat. 2019.
\newblock \href {https://doi.org/10.18653/v1/D19-1165} {Simple, scalable
  adaptation for neural machine translation}.
\newblock In \emph{Proceedings of the 2019 Conference on Empirical Methods in
  Natural Language Processing and the 9th International Joint Conference on
  Natural Language Processing (EMNLP-IJCNLP)}, pages 1538--1548, Hong Kong,
  China. Association for Computational Linguistics.

\bibitem[{Berard et~al.(2018)Berard, Besacier, Kocabiyikoglu, and
  Pietquin}]{asr-pretrain}
Alexandre Berard, Laurent Besacier, Ali~Can Kocabiyikoglu, and Olivier
  Pietquin. 2018.
\newblock \href {https://doi.org/10.1109/ICASSP.2018.8461690} {End-to-{End}
  {Automatic} {Speech} {Translation} of {Audiobooks}}.
\newblock In \emph{2018 {IEEE} {International} {Conference} on {Acoustics},
  {Speech} and {Signal} {Processing} ({ICASSP})}, pages 6224--6228, Calgary,
  AB. IEEE.

\bibitem[{Cettolo et~al.(2015)Cettolo, Niehues, St{\"u}ker, Bentivogli,
  Cattoni, and Federico}]{iwslt2015}
M.~Cettolo, J.~Niehues, S.~St{\"u}ker, L.~Bentivogli, R.~Cattoni, and Marcello
  Federico. 2015.
\newblock The iwslt 2015 evaluation campaign.
\newblock In \emph{Proceedings of the 12th {International} {Workshop} on
  {Spoken} {Language} {Translation}}.

\bibitem[{Devlin et~al.(2019)Devlin, Chang, Lee, and Toutanova}]{bert}
Jacob Devlin, Ming-Wei Chang, Kenton Lee, and Kristina Toutanova. 2019.
\newblock \href {https://doi.org/10.18653/v1/N19-1423} {{BERT}: Pre-training of
  deep bidirectional transformers for language understanding}.
\newblock In \emph{Proceedings of the 2019 Conference of the North {A}merican
  Chapter of the Association for Computational Linguistics: Human Language
  Technologies, Volume 1 (Long and Short Papers)}, pages 4171--4186,
  Minneapolis, Minnesota. Association for Computational Linguistics.

\bibitem[{Di~Gangi et~al.(2019{\natexlab{a}})Di~Gangi, Cattoni, Bentivogli,
  Negri, and Turchi}]{mustc}
Mattia~A. Di~Gangi, Roldano Cattoni, Luisa Bentivogli, Matteo Negri, and Marco
  Turchi. 2019{\natexlab{a}}.
\newblock \href {https://doi.org/10.18653/v1/N19-1202} {{M}u{ST}-{C}: a
  {M}ultilingual {S}peech {T}ranslation {C}orpus}.
\newblock In \emph{Proceedings of the 2019 Conference of the North {A}merican
  Chapter of the Association for Computational Linguistics: Human Language
  Technologies, Volume 1 (Long and Short Papers)}, pages 2012--2017,
  Minneapolis, Minnesota. Association for Computational Linguistics.

\bibitem[{Di~Gangi et~al.(2019{\natexlab{b}})Di~Gangi, Negri, Nguyen,
  Tebbifakhr, and Turchi}]{asr-pretrain-ex2}
Mattia~A. Di~Gangi, Matteo Negri, Viet~Nhat Nguyen, Amirhossein Tebbifakhr, and
  Marco Turchi. 2019{\natexlab{b}}.
\newblock \href {https://doi.org/10.5281/ZENODO.3525492} {Data {Augmentation}
  for {End}-to-{End} {Speech} {Translation}: {FBK}@{IWSLT} '19}.
\newblock In \emph{Proceedings of the 16th {International} {Workshop} on
  {Spoken} {Language} {Translation}}, Hong Kong.
\newblock Publisher: Zenodo.

\bibitem[{Di~Gangi et~al.(2019{\natexlab{c}})Di~Gangi, Negri, Nguyen,
  Tebbifakhr, and Turchi}]{fbk2019}
Mattia~A. Di~Gangi, Matteo Negri, Viet~Nhat Nguyen, Amirhossein Tebbifakhr, and
  Marco Turchi. 2019{\natexlab{c}}.
\newblock \href {https://doi.org/10.5281/zenodo.3525492} {Data augmentation for
  end-to-end speech translation: Fbk@iwslt '19}.
\newblock In \emph{Proceedings of the 16th {International} {Workshop} on
  {Spoken} {Language} {Translation}}. Zenodo.

\bibitem[{Di~Gangi et~al.(2019{\natexlab{d}})Di~Gangi, Negri, and
  Turchi}]{s-transformer}
Mattia~A. Di~Gangi, Matteo Negri, and Marco Turchi. 2019{\natexlab{d}}.
\newblock \href {https://doi.org/10.21437/Interspeech.2019-3045} {Adapting
  {Transformer} to {End}-to-{End} {Spoken} {Language} {Translation}}.
\newblock In \emph{Interspeech 2019}, pages 1133--1137. ISCA.

\bibitem[{Escolano et~al.(2020)Escolano, Costa-juss{\`a}, Fonollosa, and
  Segura}]{adapter-st}
Carlos Escolano, Marta~R Costa-juss{\`a}, Jos{\'e}~AR Fonollosa, and Carlos
  Segura. 2020.
\newblock Enabling zero-shot multilingual spoken language translation with
  language-specific encoders and decoders.
\newblock \emph{arXiv preprint arXiv:2011.01097}.

\bibitem[{Gaido et~al.(2020)Gaido, Di~Gangi, Negri, and Turchi}]{fbk2020}
Marco Gaido, Mattia~A. Di~Gangi, Matteo Negri, and Marco Turchi. 2020.
\newblock \href {https://doi.org/10.18653/v1/2020.iwslt-1.8} {End-to-end
  speech-translation with knowledge distillation: {FBK}@{IWSLT}2020}.
\newblock In \emph{Proceedings of the 17th International Conference on Spoken
  Language Translation}, pages 80--88, Online. Association for Computational
  Linguistics.

\bibitem[{Iranzo-Sánchez et~al.(2020)Iranzo-Sánchez, Silvestre-Cerdà, Jorge,
  Roselló, Giménez, Sanchis, Civera, and Juan}]{europarlst}
Javier Iranzo-Sánchez, Joan~Albert Silvestre-Cerdà, Javier Jorge, Nahuel
  Roselló, Adrià Giménez, Albert Sanchis, Jorge Civera, and Alfons Juan.
  2020.
\newblock \href {http://arxiv.org/abs/1911.03167} {Europarl-st: A multilingual
  corpus for speech translation of parliamentary debates}.

\bibitem[{{Kahn} et~al.(2020){Kahn}, {Rivière}, {Zheng}, {Kharitonov}, {Xu},
  {Mazaré}, {Karadayi}, {Liptchinsky}, {Collobert}, {Fuegen}, {Likhomanenko},
  {Synnaeve}, {Joulin}, {Mohamed}, and {Dupoux}}]{librilight}
J.~{Kahn}, M.~{Rivière}, W.~{Zheng}, E.~{Kharitonov}, Q.~{Xu}, P.~E.
  {Mazaré}, J.~{Karadayi}, V.~{Liptchinsky}, R.~{Collobert}, C.~{Fuegen},
  T.~{Likhomanenko}, G.~{Synnaeve}, A.~{Joulin}, A.~{Mohamed}, and E.~{Dupoux}.
  2020.
\newblock Libri-light: A benchmark for asr with limited or no supervision.
\newblock In \emph{ICASSP 2020 - 2020 IEEE International Conference on
  Acoustics, Speech and Signal Processing (ICASSP)}, pages 7669--7673.
\newblock \url{https://github.com/facebookresearch/libri-light}.

\bibitem[{Kharitonov et~al.(2020)Kharitonov, Rivi{\`e}re, Synnaeve, Wolf,
  Mazar{\'e}, Douze, and Dupoux}]{wavaugment2020}
Eugene Kharitonov, Morgane Rivi{\`e}re, Gabriel Synnaeve, Lior Wolf,
  Pierre-Emmanuel Mazar{\'e}, Matthijs Douze, and Emmanuel Dupoux. 2020.
\newblock Data augmenting contrastive learning of speech representations in the
  time domain.
\newblock \emph{arXiv preprint arXiv:2007.00991}.

\bibitem[{Kingma and Ba(2017)}]{adam}
Diederik~P. Kingma and Jimmy Ba. 2017.
\newblock \href {http://arxiv.org/abs/1412.6980} {Adam: A method for stochastic
  optimization}.

\bibitem[{Lakumarapu et~al.(2020)Lakumarapu, Lee, Indurthi, Han, Zaidi, and
  Kim}]{bhanss2020}
Nikhil~Kumar Lakumarapu, Beomseok Lee, Sathish~Reddy Indurthi, Hou~Jeung Han,
  Mohd~Abbas Zaidi, and Sangha Kim. 2020.
\newblock \href {https://doi.org/10.18653/v1/2020.iwslt-1.7} {End-to-end
  offline speech translation system for {IWSLT} 2020 using modality agnostic
  meta-learning}.
\newblock In \emph{Proceedings of the 17th International Conference on Spoken
  Language Translation}, pages 73--79, Online. Association for Computational
  Linguistics.

\bibitem[{Lewis et~al.(2020)Lewis, Liu, Goyal, Ghazvininejad, Mohamed, Levy,
  Stoyanov, and Zettlemoyer}]{BART}
Mike Lewis, Yinhan Liu, Naman Goyal, Marjan Ghazvininejad, Abdelrahman Mohamed,
  Omer Levy, Veselin Stoyanov, and Luke Zettlemoyer. 2020.
\newblock \href {https://doi.org/10.18653/v1/2020.acl-main.703} {{BART}:
  Denoising sequence-to-sequence pre-training for natural language generation,
  translation, and comprehension}.
\newblock In \emph{Proceedings of the 58th Annual Meeting of the Association
  for Computational Linguistics}, pages 7871--7880, Online. Association for
  Computational Linguistics.

\bibitem[{Li et~al.(2021)Li, Wang, Tang, Tran, Tang, Pino, Baevski, Conneau,
  and Auli}]{lna}
Xian Li, Changhan Wang, Yun Tang, Chau Tran, Yuqing Tang, Juan Pino, Alexei
  Baevski, Alexis Conneau, and Michael Auli. 2021.
\newblock \href {http://arxiv.org/abs/2010.12829} {Multilingual speech
  translation with efficient finetuning of pretrained models}.

\bibitem[{Liu et~al.(2020{\natexlab{a}})Liu, Gu, Goyal, Li, Edunov,
  Ghazvininejad, Lewis, and Zettlemoyer}]{mBART}
Yinhan Liu, Jiatao Gu, Naman Goyal, Xian Li, Sergey Edunov, Marjan
  Ghazvininejad, Mike Lewis, and Luke Zettlemoyer. 2020{\natexlab{a}}.
\newblock Multilingual denoising pre-training for neural machine translation.
\newblock \emph{Transactions of the Association for Computational Linguistics},
  8:726--742.

\bibitem[{Liu et~al.(2020{\natexlab{b}})Liu, Gu, Goyal, Li, Edunov,
  Ghazvininejad, Lewis, and Zettlemoyer}]{mBART25}
Yinhan Liu, Jiatao Gu, Naman Goyal, Xian Li, Sergey Edunov, Marjan
  Ghazvininejad, Mike Lewis, and Luke Zettlemoyer. 2020{\natexlab{b}}.
\newblock \href {http://arxiv.org/abs/2001.08210} {Multilingual denoising
  pre-training for neural machine translation}.

\bibitem[{Meignier and Merlin(2010)}]{lium}
Sylvain Meignier and Teva Merlin. 2010.
\newblock Lium spkdiarization: an open source toolkit for diarization.
\newblock In \emph{CMU SPUD Workshop}.

\bibitem[{Niehues et~al.(2019)Niehues, Cattoni, St{\"u}ker, Negri, Turchi,
  Salesky, Sanabria, Barrault, Specia, and Federico}]{iwslt2019}
J.~Niehues, R.~Cattoni, S.~St{\"u}ker, M.~Negri, M.~Turchi, Elizabeth Salesky,
  Ramon Sanabria, Lo{\"i}c Barrault, Lucia Specia, and Marcello Federico. 2019.
\newblock The iwslt 2019 evaluation campaign.
\newblock In \emph{Proceedings of the 16th {International} {Workshop} on
  {Spoken} {Language} {Translation}}.

\bibitem[{Panayotov et~al.(2015)Panayotov, Chen, Povey, and
  Khudanpur}]{librispeech}
Vassil Panayotov, Guoguo Chen, Daniel Povey, and Sanjeev Khudanpur. 2015.
\newblock \href {https://doi.org/10.1109/ICASSP.2015.7178964} {Librispeech: An
  asr corpus based on public domain audio books}.
\newblock In \emph{2015 IEEE International Conference on Acoustics, Speech and
  Signal Processing (ICASSP)}, pages 5206--5210.

\bibitem[{Park et~al.(2019)Park, Chan, Zhang, Chiu, Zoph, Cubuk, and
  Le}]{specaugment}
Daniel~S. Park, William Chan, Yu~Zhang, Chung-Cheng Chiu, Barret Zoph, Ekin~D.
  Cubuk, and Quoc~V. Le. 2019.
\newblock \href {https://doi.org/10.21437/Interspeech.2019-2680}
  {{SpecAugment}: {A} {Simple} {Data} {Augmentation} {Method} for {Automatic}
  {Speech} {Recognition}}.
\newblock In \emph{Interspeech 2019}, pages 2613--2617. ISCA.

\bibitem[{Pino et~al.(2019)Pino, Puzon, Gu, Ma, McCarthy, and
  Gopinath}]{asr-pretrain-ex1}
Juan Pino, Liezl Puzon, Jiatao Gu, Xutai Ma, Arya~D. McCarthy, and Deepak
  Gopinath. 2019.
\newblock \href {https://doi.org/10.5281/ZENODO.3525032} {Harnessing {Indirect}
  {Training} {Data} for {End}-to-{End} {Automatic} {Speech} {Translation}:
  {Tricks} of the {Trade}}.
\newblock In \emph{Proceedings of the 16th {International} {Workshop} on
  {Spoken} {Language} {Translation}}, Hong Kong.
\newblock Publisher: Zenodo.

\bibitem[{Post(2018)}]{sacrebleu}
Matt Post. 2018.
\newblock \href {https://doi.org/10.18653/v1/W18-6319} {A call for clarity in
  reporting {BLEU} scores}.
\newblock In \emph{Proceedings of the Third Conference on Machine Translation:
  Research Papers}, pages 186--191, Brussels, Belgium. Association for
  Computational Linguistics.

\bibitem[{Potapczyk and Przybysz(2020)}]{best_iwslt2020}
Tomasz Potapczyk and Pawel Przybysz. 2020.
\newblock \href {https://doi.org/10.18653/v1/2020.iwslt-1.9} {{SRPOL}{'}s
  system for the {IWSLT} 2020 end-to-end speech translation task}.
\newblock In \emph{Proceedings of the 17th International Conference on Spoken
  Language Translation}, pages 89--94, Online. Association for Computational
  Linguistics.

\bibitem[{Potapczyk et~al.(2019)Potapczyk, Przybysz, Chochowski, and
  Szumac}]{srpol2019}
Tomasz Potapczyk, Pawel Przybysz, Marcin Chochowski, and Artur Szumac. 2019.
\newblock \href {https://doi.org/10.5281/zenodo.3525498} {Samsung's system for
  the iwslt 2019 end-to-end speech translation task}.
\newblock In \emph{Proceedings of the 16th {International} {Workshop} on
  {Spoken} {Language} {Translation}}. Zenodo.

\bibitem[{Tang et~al.(2020)Tang, Tran, Li, Chen, Goyal, Chaudhary, Gu, and
  Fan}]{mBART50}
Yuqing Tang, Chau Tran, Xian Li, Peng-Jen Chen, Naman Goyal, Vishrav Chaudhary,
  Jiatao Gu, and Angela Fan. 2020.
\newblock Multilingual translation with extensible multilingual pretraining and
  finetuning.
\newblock \emph{arXiv preprint arXiv:2008.00401}.

\bibitem[{Vaswani et~al.(2017)Vaswani, Shazeer, Parmar, Uszkoreit, Jones,
  Gomez, Kaiser, and Polosukhin}]{transformer}
Ashish Vaswani, Noam Shazeer, Niki Parmar, Jakob Uszkoreit, Llion Jones,
  Aidan~N Gomez, \L~ukasz Kaiser, and Illia Polosukhin. 2017.
\newblock \href
  {https://proceedings.neurips.cc/paper/2017/file/3f5ee243547dee91fbd053c1c4a845aa-Paper.pdf}
  {Attention is all you need}.
\newblock In \emph{Advances in Neural Information Processing Systems},
  volume~30. Curran Associates, Inc.

\bibitem[{Wang et~al.(2020{\natexlab{a}})Wang, Tang, Ma, Wu, Okhonko, and
  Pino}]{wang-etal-2020-fairseq}
Changhan Wang, Yun Tang, Xutai Ma, Anne Wu, Dmytro Okhonko, and Juan Pino.
  2020{\natexlab{a}}.
\newblock \href {https://www.aclweb.org/anthology/2020.aacl-demo.6} {Fairseq
  {S}2{T}: Fast speech-to-text modeling with fairseq}.
\newblock In \emph{Proceedings of the 1st Conference of the Asia-Pacific
  Chapter of the Association for Computational Linguistics and the 10th
  International Joint Conference on Natural Language Processing: System
  Demonstrations}, pages 33--39, Suzhou, China. Association for Computational
  Linguistics.

\bibitem[{Wang et~al.(2020{\natexlab{b}})Wang, Wu, and Pino}]{covost}
Changhan Wang, Anne Wu, and Juan Pino. 2020{\natexlab{b}}.
\newblock \href {http://arxiv.org/abs/2007.10310} {Covost 2: A massively
  multilingual speech-to-text translation corpus}.

\bibitem[{Weiss et~al.(2017)Weiss, Chorowski, Jaitly, Wu, and
  Chen}]{end2end-st}
Ron~J. Weiss, Jan Chorowski, Navdeep Jaitly, Yonghui Wu, and Zhifeng Chen.
  2017.
\newblock \href {https://doi.org/10.21437/Interspeech.2017-503}
  {Sequence-to-{Sequence} {Models} {Can} {Directly} {Translate} {Foreign}
  {Speech}}.
\newblock In \emph{Interspeech 2017}, pages 2625--2629. ISCA.

\bibitem[{Wolf et~al.(2020)Wolf, Debut, Sanh, Chaumond, Delangue, Moi, Cistac,
  Rault, Louf, Funtowicz, Davison, Shleifer, von Platen, Ma, Jernite, Plu, Xu,
  Le~Scao, Gugger, Drame, Lhoest, and Rush}]{transformers}
Thomas Wolf, Lysandre Debut, Victor Sanh, Julien Chaumond, Clement Delangue,
  Anthony Moi, Pierric Cistac, Tim Rault, Remi Louf, Morgan Funtowicz, Joe
  Davison, Sam Shleifer, Patrick von Platen, Clara Ma, Yacine Jernite, Julien
  Plu, Canwen Xu, Teven Le~Scao, Sylvain Gugger, Mariama Drame, Quentin Lhoest,
  and Alexander Rush. 2020.
\newblock \href {https://doi.org/10.18653/v1/2020.emnlp-demos.6} {Transformers:
  State-of-the-art natural language processing}.
\newblock In \emph{Proceedings of the 2020 Conference on Empirical Methods in
  Natural Language Processing: System Demonstrations}, pages 38--45, Online.
  Association for Computational Linguistics.

\bibitem[{Xu et~al.(2020)Xu, Baevski, Likhomanenko, Tomasello, Conneau,
  Collobert, Synnaeve, and Auli}]{wav2vec-selftraining}
Qiantong Xu, Alexei Baevski, Tatiana Likhomanenko, Paden Tomasello, Alexis
  Conneau, Ronan Collobert, Gabriel Synnaeve, and Michael Auli. 2020.
\newblock Self-training and pre-training are complementary for speech
  recognition.
\newblock \emph{arXiv preprint arXiv:2010.11430}.

\end{thebibliography}


\end{document}